# Medical Diagnosis with a Novel SVM-CoDOA Based Hybrid Approach

*M. Hanefi Calp*
Karadeniz Technical University
Faculty of Economics & Administrative Sciences
Department of Management Information Systems
61080, Trabzon, Turkey
Phone: +90 462 377 3000
mhcalp@ktu.edu.tr

**Abstract**

Machine Learning is an important sub-field of the Artificial Intelligence and it has been become a very critical task to train Machine Learning techniques via effective method or techniques. Recently, researchers try to use alternative techniques to improve ability of Machine Learning techniques. Moving from the explanations, objective of this study is to introduce a novel SVM-CoDOA (Cognitive Development Optimization Algorithm trained Support Vector Machines) system for general medical diagnosis. In detail, the system consists of a SVM, which is trained by CoDOA, a newly developed optimization algorithm. As it is known, use of optimization algorithms is an essential task to train and improve Machine Learning techniques. In this sense, the study has provided a medical diagnosis oriented problem scope in order to show effectiveness of the SVM-CoDOA hybrid formation.

## 1. Introduction

Artificial Intelligence has taken many steps since its first introduction to the scientific community. Including many different problem and application types like classification, recognition, control, diagnosis, and prediction (Russell, 2016; Ertel, 2018; Nabiyev, 2005), Artificial Intelligence has become an important scientific field for our life. Because of its successful outputs in different fields, it has already become a multidisciplinary science and a remarkable research interest in this manner has risen accordingly. Today, we can see the associated literature of Artificial Intelligence is very powerful and two important sub-fields: Machine Learning, and intelligent optimization have the most reputable role in this manner. Both these sub-fields form the exact learning intelligent system infrastructure via different solution approaches logically and mathematically advanced and inspired from especially the nature or swarms in the real life (Michalski et. al., 2013; Alpaydin, 2009; Eberhart et. al., 2001; Yang, 2010). While Machine Learning has an active role in the core of the Artificial Intelligence field, with its mechanism to learn from samples or experiences for a trained intelligent system, intelligent optimization just uses certain algorithms – techniques to deal with optimization tasks. But before training of Machine Learning techniques is a typical optimization, intelligent optimization algorithms – techniques are widely used for training Machine Learning techniques. In this way, different types of hybrid systems are developed for better solution approaches (Hojjat, 2018; Bahrami et. al., 2017; Akbal, 2018; Kumar et. al., 2018; Frazzon et. al., 2018).

Artificial Intelligence and the recent research interest: using hybrid system are widely applied in research studies within different fields. The more the objective research becomes important, the more use of hybrid system has become popular. As associated with that, an important application scope of hybrid systems has become medical diagnosis. Because it is a vital factor to have accurate diagnosis results for certain diseases, researchers are currently focused on developing alternative and effective intelligent solutions for medical diagnosis. In the associated literature, it is possible to see many different examples of using Artificial Intelligence and even hybrid systems for medical diagnosis purposes (Kononenko, 2001; Dilsizian & Siegel, 2014; Amato et. al., 2013; Choi et. al., 2017; Malav et. al., 2017; Hassanien et. al., 2014; Kumar et. al., 2015; Vasant, 2018; Cankaya et. al., 2018; Karakoc, 2018). Since the future is connected with the technological advantages of Artificial Intelligence, it becomes a remarkable research tasks to search for alternative systems performing good diagnosis performances for medical data.

Moving from the explanations so far, objective of this study is to introduce a novel SVM-CoDOA (Cognitive Development Optimization Algorithm trained Support Vector Machines) system for general medical diagnosis. In detail, the system consists of a SVM, which is trained by CoDOA, a newly developed optimization algorithm. As it is known, use of optimization algorithms is an essential task to train and improve Machine Learning techniques. In this sense, the work provides a medical diagnosis oriented problem scope in order to show effectiveness of the SVM-CoDOA hybrid formation.

This study is a general research report for the performed study. So, it is aimed firstly to express some about the employed techniques for the whole SVM-CoDOA hybrid system, which was applied in the scope of medical diagnosis topic. In detail, it is also important to have information about how the SVM is trained by the CoDOA so that the necessary explanations are provided accordingly. The system developed in this study was used for some known medical data in order to have accurate information about its performance and diagnosis ability.

## 2. Methods and the Problem Solution

It is important to have information about what is lying behind the hybrid system considered here and how the objective medical diagnosis problem has been solved in this context. The following sub-sections are devoted to that. The techniques – algorithms utilized in the study have been: SVM, CoDOA, Genetic Algorithm (GA), Differential Evolution algorithm (DE), Clonal Selection Algorithm (CSA), and Particle Swarm Optimization algorithm (PSO) respectively. These techniques are briefly described below.

### 2.1. Support Vector Machines (SVM)

As introduced in 1979 by Vapnik and Lerner, Support Vector Machines (SVM) is used for especially classification and regression tasks (Comak et. al., 2007; Vapnik et. al., 1997; Santhanam, 2015). It briefly focuses on finding the optimally separating hyper-plane classifying the target data. In this context, the margin(s) regarding the classes-groups are tried to be maximized, which means a typical optimization task. Support vectors are called here as typical subset of data instances associated with the term: hyper-plane and the distance between nearest support vector and the hyper-plane is a margin. In detail, SVM actually deals with both regression and classification problems by using its two typical variations: linear SVM and non-linear SVM. Linear one is for separating the data with linear decision boundary (especially for two classes-groups) and the non-linear one is for separating the data thanks to nonlinear decision boundary (Santhanam, 2015; Tamura & Tanno, 2009). Objective of a SVM is to find the optimum separating hyper-plane, which can classify data points and divide them with classification points. SVM in this manner aims to determine the state in which the distance between different (generally two) classes is at the maximum level. Figure 1 shows an example for support vectors and the maximum / optimum separating margin hyper-plane (Hepworth et. al., 2012).

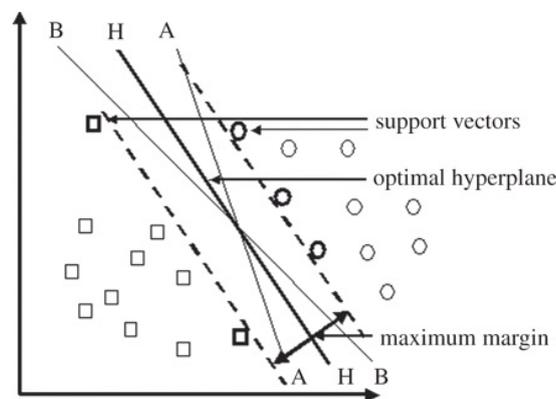

*Figure 1. SVM and the related components (Hepworth et. al., 2012)*

Linear SVM corresponds to classifying the data points with two classes. But in many applications, the objective data is indistinguishable with a linear way so that issue is solved thanks to a higher dimensional space approach. The solution approach here is based on that the non-linear support vector regarding the original input space is adjusted in a higher dimensional feature space, which can be divided into the training data. As called as non-linear SVM, that type of the technique benefits from kernel function in order to transform m-dimensional data into a new data set, which corresponds to a higher dimension classification, which is still linear. At this point, choosing the appropriate kernel function is important because it is connected with the exact performance shown. Some widely used kernel functions in non-linear SVM are as follows (Cankaya et. al., 2018):

Linear function: $K(x_i, x_j) = x_i^T x_j$ (1)

Polynomial function: $K(x_i, x_j) = (1 + x_i^T x_j)^p$ (2)

Gaussian (RBF) function: $K(x_i, x_j) = e^{\frac{\|x_i - x_j\|^2}{2\sigma^2}}$ (3)

### 2.2. A Recent Optimization Technique: Cognitive Development Optimization Algorithm

Cognitive Development Optimization Algorithm (CoDOA) is an Artificial Intelligence based optimization technique, which can be used for optimization tasks. That optimization algorithm is briefly inspired from the Piaget's Theory on Cognitive Development. In detail, the cognitive development corresponds to our natural development as expressed by Piaget and we experience some typical process like social interaction, maturation, or balancing during learning something in this manner (Kose, 2017; Piaget, 1964; Piaget, 1973; Singer & Revenson, 1997). Similarly, the CoDOA includes some certain solution stages as: Starting, Socialization, Maturation, Rationalizing, and Balancing. The stages include mathematical and logical operations around $N$ particles taking role in the objective optimization solution space (Kose, 2017).

The steps of the CoDOA are briefly as follows (Kose, 2017):
- **1st Step (Starting):** Adjust initial parameters like $N$: number of particles, *ir:* initial interactivity rate, *r:* rationality rate, *ml:* maturity limit, *ex:* experience for each particle, max. and min. value for *ir*.

- **2nd Step:** Spread particles randomly to the solution space. Calculate fitness. Renew *ir* of the best particle with the following equation and increase its *ex* value by 1 (b is for the best, p is for the particle):

  b_p_ir_(new) = b_p_ir_(current) + (rand. * b_p_ir_(current))  (4)

- **3rd Step:** Repeat the following steps till the stop criterion is met:

  **3.1st Step (Socialization):** Decrease *ex* the particles, whose fitness is equal to or above average fitness by 1 (considering the problem is minimization). In addition, increase *ex* of the particles in contrast situation, by 1. (Consideration the problem is minimization). Also, renew *ir* of these particles via the following:

  p_j_ir_(new) = p_j_ir_(current) + (rand. * p_j_ir_(current))  (5)

  **3.2nd Step:** Renew *ir* of all particles via:

  p_i_ir_(new) = rand. * p_i_ir_(current)  (6)

  **3.3th Step 3.3:** Renew position of each particle (except from the best one) via the following:

  p_i_pos._(new) = p_i_pos._(current) + (rand. * (p_i_ir_(current) * (global_best_pos. – p_i_pos._(current))))  (7)

  **3.4th Step:** Calculate fitness. Renew *ir* of the best particle with a random value and increase its *ex* by 1:

  b_p_ir_(new) = b_p_ir_(current) + (rand. * b_p_ir_(current))  (8)

**3.5th Step (Maturation):** Renew *ir* of the particle, whose *ex* is equal to or under the *ml*, with the following:

$$p_{j\_ir\_}(new) = p_{j\_ir\_}(current) + (rand. * p_{j\_ir\_}(current)) \tag{9}$$

Calculate fitness and update *ir* of the best particle and increase its *ex* by 1:

$$b\_p\_ir\_(new) = b\_p\_ir\_(current) + (rand. * b\_p\_ir\_(current)) \tag{10}$$

**3.6th Step (Rationalizing):** Renew *ir* and positions of the particles, whose *ex* is under 0:

$$p_{j\_ir\_}(new) = p_{j\_ir\_}(current) + (rand. * (b\_p\_ir\_(current) / p_{j\_ir\_}(current))) \tag{11}$$

$$p_{i\_pos.\_}(new) = p_{i\_pos.\_}(current) + (rand. * (p_{i\_ir\_}(current) * (global\_best\_pos. - p_{i\_pos.\_}(current)))) \tag{12}$$

Renew *ir* of the particles, whose *ex* is equal to or above 0, and repeat that *r* times, via the following:

$$p_{j\_ir\_}(new) = p_{j\_ir\_}(current) + (rand. * (b\_p\_ir\_(current) / p_{j\_ir\_}(current))) \tag{13}$$

**3.7th Step 3.7 (Balancing):** Renew *ir* of all particles by using the following:

$$p_{i\_ir\_}(new) = rand. * p_{i\_ir\_}(current) \tag{14}$$

Calculate fitness. Renew *ir* of the best particle, increase its *ex* by 1 and realize in-system optimization for advanced problems:

$$b\_p\_ir\_(new) = b\_p\_ir\_(current) + (rand. * b\_p\_ir\_(current)) \tag{15}$$

Return to the 3.1st Step, if the stopping criteria is not met yet.

- **4th Step:** End of the optimization process. The result(s) are the best / optimum value(s).

By following the related algorithmic steps, the CoDOA is able to deal with optimization tasks. In this context, that mechanism was used for training purposes of SVM, as explained under the following paragraphs.

**2.3. Genetic Algorithm (GA)**

GA is the first artificial intelligence technique inspired by basic elements and phenomena of the Theory of Evolution. GA is generally based on subjecting a variety of genetic manipulations according to the fitness values of individuals which are represented as chromosomes. In other definition, GA determines the fitness value of each individual, randomly selects a number of pairs of individuals, generates two new individuals (offspring solutions) from each selected pair of parent solutions, and uses a two-point crossover operator for this. That is, according to the results of the initial population, processes such as crossing and mutation are carried out. New individuals are obtained through relatively better individuals. The process required to create each new generation continues until the desired result is obtained or a certain stop criterion is reached. In summary, GAs produce resolutions to optimization problems utilizing methods of natural evolution (inheritance, mutation, selection, and crossover) (Kramer, 2017; Holand, 1992; Calp & Akcayol, 2018; John & Krishnakumar, 2017; Kadri & Boctor, 2018; Dener & Calp, 2018). In Table 1, general framework of GA was given.

Table 1. The Framework of GA

| |
|---|
| **Step 1:** Creation of population *(with n individuals, according to target problem).* *Iterative steps for each individual and each purpose function size:* |
| **Step 2:** Calculation of fitness function value. |
| **Step 3:** Selection of individuals who will enter reproductive process. |
| **Step 4:** Crossing of selected individuals. |
| **Step 5:** Mutation of some individuals. |
| **Step 6:** Obtaining of optimal solution *(It is the global best position and value(s) at end of the iterative process).* |

### 2.4. Differential Evolution Algorithm (DE)

Differential Evolution (DE) is a parallel direct search method. The DE algorithm is a population-based algorithm using similar with genetic algorithms operators; crossover, mutation and selection. The initial vector populations chosen randomly and should cover the entire parameter space. The main difference in constructing better solutions is that genetic algorithms rely on crossover while DE relies on mutation operation. The algorithm uses mutation operation as a search mechanism and selection operation to direct the search toward the prospective regions in the search space. The DE algorithm also uses a non-uniform crossover that can take child vector parameters from one parent more often than it does from others. By using the components of the existing population members to construct trial vectors, the recombination (crossover) operator efficiently shuffles information about successful combinations, enabling the search for a better solution space (Storn & Price, 1997; Fleetwodd, 2004; Karaboga & Okdem, 2004; Mallipeddi et. al., 2011). In Table 2, general framework of DE was given.

Table 2. The Framework of DE

**Step 1:** Initialization *(Randomly initialize the population of an individual)*
**Step 2:** Evaluation *(Evaluation of the objective values of all individuals)*
 *Repeat Until termination criteria are met*
**Step 3:** Mutation *(Generating of a donor vector)*
**Step 4:** Recombination *(crossover)*
**Step 5:** Evaluation *(Evaluating of the objective values of the trial vectors)*
**Step 6:** Selection *(Performing of a selection operation between each individual and its corresponding trial vector in order to generate the new individual for the next generation)*

### 2.5. Clonal Selection Algorithm (CSA)

CSA is a class of algorithms inspired by the clonal selection of acquired immunity. These algorithms focus on the Darwinian attributes of the theory, where selection is inspired by the affinity of antigen–antibody interactions, reproduction is inspired by cell division, and variation is inspired by somatic hyper-mutation. CSA was modeled by being inspired by the principle of biological clonal selection. The purpose of the clonal selection principle is to provide the antibody diversity that can fight against the antigens. Whenever a new antigen is encountered, the immune network is updated according to these antigens, thereby increasing the identifiability of the antigens (Xu et. al., 2018; Bernardino et. al., 2011; Yavuz et. al., 2018; Ulutas & Kulturel-Konak, 2011). In Table 3, general framework of CSA was given.

Table 3. The Framework of CSA

**Step 1:** Initialization *(Randomly initialize a population N of antibodies)*
 *Repeat steps 2–5 until termination criterion is met*
**Step 2:** Evaluation *(Determine the affinity of each antibody)*
**Step 3:** Selection and Cloning *(Select a number (n) of the highest affinity antibodies and generate clones independently and proportionally to their affinities)*
**Step 4:** Hyper-Mutation *(Generating matured clones. The higher the affinity, the smaller the mutation rate)*
**Step 5:** Clone Evaluation and Reselection *(Determine the affinity of the matured clones in relation to antigen. Select the antibody with the highest affinity from the matured clones and form the new population N.)*

### 2.6. Particle Swarm Optimization Algorithm (PSO)

Particle Swarm Optimization (PSO) is one of the most fundamental of intelligent optimization algorithms. In the algorithm, iterative steps are taken in order to find the optimum of the N particles scattered in the solution space and at this point parameters such as position and velocity are the determinants of general movements. The particle positions obtained as a result of the movements are used to calculate the fitness function (s) of the problem within the scope of the variables they correspond to. The best personal position (optimal value) obtained by each particle so far during position changes and the global position value (position that provides the global optimum value) so far in the solution process are taken into account. The particles are trying to reach a solution by following the particle which provides the global optimum under the influence of these values. In other words and summary, PSO is an algorithm developed in general by looking for food search movements in flocks of animals such as birds and fish in nature. The PSO algorithm operates on a population (swarm) of candidate solutions (particles). The algorithm begins with particles initialized to random positions. Each iteration updates every particle's velocity. After all particles are updated, the global best is updated from the swarm's current set of personal bests (Kennedy, 2011; Clerc, 2010; Butcher et. al., 2018; Eberhart & Kennedy, 1995; Shi & Eberhart, 1998). In Table 4, general framework of CSA was given.

Table 4. The Framework of PSO

**Step 1:** Setup: *Randomly dispense the N particle in the solution space and set the initial Speed (v: velocity) value for each particle. Make the starting position of each particle at the same time the best personal position. Assign the values of the algorithm parameters. Make arrangements for the problem to be solved.*
*Iterative steps (for each particle and each fitness function size):*
**Step 2:** Calculating of the fitness function value according to the position of the particle.
**Step 3:** Updating of position of the particle (if the location of the snippet is better than your personal best location so far).
**Step 4:** Updating of speed of the particle.
**Step 5:** Updating of location of the particle with the current speed.
**Step 6:** Obtaining of optimum solution *(At the end of the iterative process, the optimum solution is the global best position and value (s)).*

### 2.7. Medical Diagnosis with the Novel SVM-CoDOA

In this study, it is aimed to perform medical diagnosis by using the appropriate data, which is suitable to be used for classification purposes. In detail, the SVM is trained with the CoDOA in order to achieve the appropriate, trained SVM, which can classify the newly encountered data for the accurate diagnosis. In detail, some remarkable points of the introduced problem solution for the medical diagnosis are as follows (Figure 2):

1. The main objective is to find the optimum sigma ($\sigma$) parameter of the Gaussian (RBF) kernel function of the non-linear SVM.
2. In order to find the optimum sigma ($\sigma$), each particle of the CoDOA corresponds to that parameter so that they can move through the optimization process in order to find the optimum value.
3. For an objective medical diagnosis problem, CoDOA is run according to a specific iteration number (stopping criterion). In each iteration turn, the determined sigma ($\sigma$) values are used for the diagnosis over the training data. In this context, the following diagnosis accuracy calculation is used to determine the best particle in the related iteration and renew the particle – algorithm parameters by considering it:

$$100 * \frac{True\ Diagnosis}{True\ Diagnosis + Fa\quad Diagnosis} \tag{16}$$

4. At the end of the optimization process regarding the CoDOA, the best particle has the optimum sigma ($\sigma$) value allowing the non-linear SVM to perform the best classification – diagnosis for the objective medical diagnosis problem.

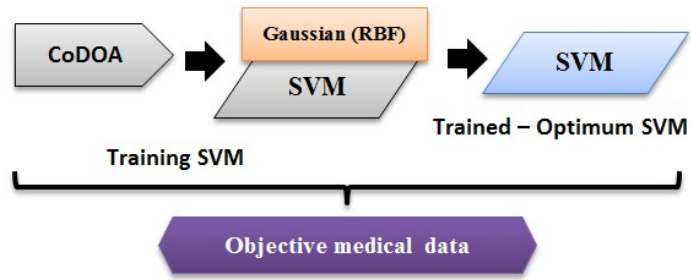

*Figure 2. SVM-CoDOA system for medical diagnosis*

### 3. Medical Diagnosis Applications

In order to see its performance on medical data, SVM-CoDOA system was applied in three different medical data obtained from the UCI Repository (UCI, 2018). As the Thyroid data set, the first set includes 2800 training data, and 972 test data respectively with 21 attributes considered. The second data set: Hepatitis data set includes 155 data (as separated into 100 training data, 55 test data in this study), with 19 attributes. Finally, the third data set: Chronic Kidney Disease data set includes 400 data (as separated into 295 training data, 105 test data in this study), with 25 attributes. For the CoDOA, default parameters were set as *ir*: 0.50, max. interactivity: 10, *ml*:3, and rationality rate: 2 with a total of 90 particles (*N*) run in a total of 5000 iterations, as suggested by (Kose et. al., 2016).

Considering the diagnosis accuracy calculation expressed in the Equation 16, Table 5 provides findings obtained with the SVM-CoDOA system for three separate medical data sets.

Table 5. Findings obtained with the SVM-CoDOA system for three separate medical data sets

| Data Set | Test Data | True Diagnosis | False Diagnosis | Accuracy (%) |
|---|---|---|---|---|
| Thyroid Disease | 972 | 939 | 33 | 96.60 |
| Hepatitis Disease | 55 | 49 | 6 | 89.10 |
| Chronic Kidney Disease | 105 | 91 | 14 | 86.67 |

In addition to the self-performance evaluation, the SVM-CoDOA system was compared with also four alternative systems formed with different intelligent optimization techniques – algorithms (Genetic Algorithm (GA), Differential Evolution algorithm (DE), Clonal Selection Algorithm (CSA), and Particle Swarm Optimization algorithm (PSO)) but same Gaussian (RBF) kernel function based non-linear SVM. Default parameters of these algorithms from were used for running same 5000-iteration optimization process with 90 particles. For accurate findings, the whole systems including SVM-CoDOA were run 50 times and average accuracy values for three medical diagnosis data sets were considered. Table 6 provides findings of true and false diagnosis for all systems and the objective data sets (Best values are in bold).

Table 6. Average findings obtained with all alternative hybrid systems, after 50 independent runs

| Data Set | SVM-CoDOA | | SVM-GA | | SVM-DE | | SVM-CSA | | SVM-PSO | |
|---|---|---|---|---|---|---|---|---|---|---|
| | TD* | FD* | TD* | FD* | TD* | FD* | TD* | FD* | TD* | FD* |
| Thyroid Disease | **934** | **38** | 860 | 112 | 906 | 66 | 883 | 89 | 897 | 75 |
| Hepatitis Disease | 47 | 8 | 39 | 16 | **49** | **6** | 44 | 11 | 43 | 12 |
| Chronic Kidney Disease | **88** | **17** | 79 | 26 | 84 | 21 | 85 | 20 | 81 | 24 |
| * TD: True Diagnosis / FD: False Diagnosis / Best values are in bold. | | | | | | | | | | |

Considering the findings, Figure 3 shows graphics of average accuracy values for each system and the objective medical diagnosis data set.

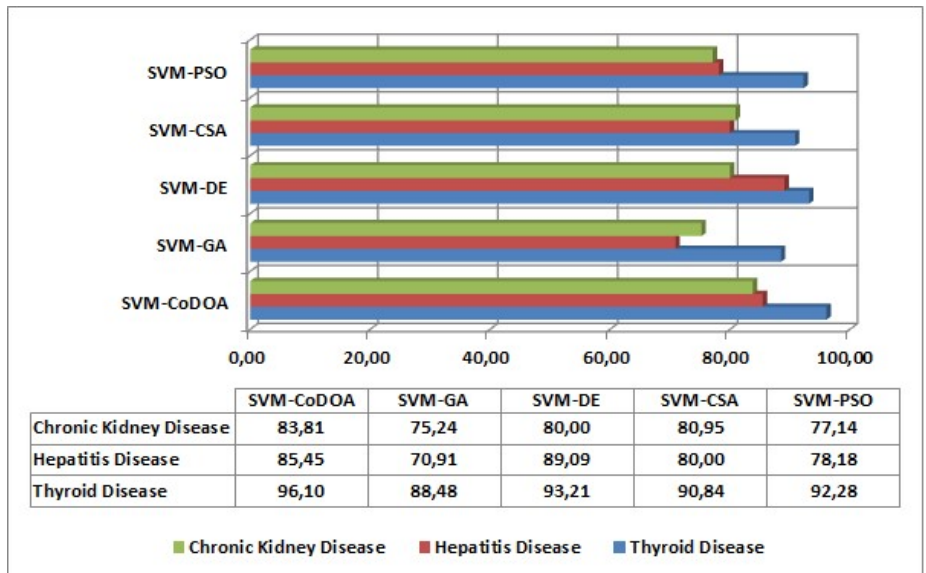

*Figure 3. Average accuracy of different SVM based hybrid systems for objective medical diagnosis data sets*

As it can be seen from the findings, the SVM-CoDOA system developed in this study is successful enough to perform accurate medical diagnosis for even different diseases. In detail, the system is not only successful in its own solution approach but also better in general than some other alternative SVM based hybrid systems formed with different intelligent optimization techniques – algorithms. Although the SVM-DE is the best performing for the Hepatitis disease data set, SVM-CoDOA provides very near results to SVM-DE in terms of diagnosis.

**4. Conclusions and Future Work**

In this study, a SVM-CoDOA system was introduced for medical diagnosis problem. In detail, a recent intelligent optimization (Artificial Intelligence based optimization) algorithm: Cognitive Development Optimization Algorithm (CoDOA) was used for training a non-linear Support Vector Machines (SVM) for better classification, which means also better diagnosis. Briefly, CoDOA particles were employed for finding the optimum sigma (σ) parameter of the Gaussian kernel function of the SVM so that it can perform an accurate diagnosis over the objective medical data. In the application works, the developed SVM-CoDOA was applied in some different medical data and the obtained results showed that the system is effective enough in diagnosis operations. Further, the formed system is also better than four different SVM based systems formed via alternative intelligent optimization techniques – algorithms.

In addition to the performed research, there are also some alternative future works planned by the author. Briefly, the system will be used for alternative medical data and the effect of different parameters will be evaluated. Finally, there will also some efforts to use CoDOA for different parameter optimization of the non-linear SVM.

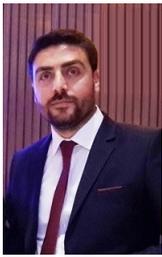

Dr. **Hanefi M. Calp** received his PhD degree in 2018 from the department of Management Information Systems at Gazi University, one of the most prestigious universities in Turkey. He works as an Assistant Professor in the department of Management Information Systems of the Faculty of Economics & Administrative Sciences of the Karadeniz Technical University. His research interest includes Management Information Systems, Artifical Neural Networks, Expert Systems, Fuzzy Logic, Risk Management, Risk Analysis, Human-Computer Interaction, Technology Management and Project Management.